\documentclass{article}
\usepackage{spconf,amsmath,graphicx,hyperref}
\usepackage{algorithm}
\usepackage{algorithmic}
\usepackage{graphicx}
\usepackage{booktabs}  
\usepackage{multirow}  
\usepackage{graphicx}  
\usepackage{siunitx}   
\usepackage{amsmath} 
\usepackage{xspace}
\usepackage[table,xcdraw]{xcolor}
\definecolor{lightblue}{rgb}{0.93,0.95,1.0}

\title{MindCine: Multimodal EEG-to-Video Reconstruction with\\ Large-Scale Pretrained Models}
\name{Tian-Yi Zhou$^*$\thanks{$^*$Equal contribution.\texttt{\{ty\_zhou, haogram\_sjtu\}@sjtu.edu.cn}} \qquad Xuan-Hao Liu$^*$ \qquad Bao-Liang Lu \qquad Wei-Long Zheng$^\dagger$\thanks{$^\dagger$Corresponding author: \texttt{weilong@sjtu.edu.cn}}}
\address{School of Computer Science, Shanghai Jiao Tong University, 800 Dongchuan Road, Shanghai, China}

\begin{document}
\ninept 
\maketitle
\begin{abstract}
Reconstructing human dynamic visual perception from electroencephalography (EEG) signals is of great research significance since EEG’s non-invasiveness and high temporal resolution.
However, EEG-to-video reconstruction remains challenging due to: \textbf{\textit{1) Single Modality}} existing studies solely align EEG signals with the text modality, which ignores other modalities and are prone to suffer from overfitting problems; \textbf{\textit{2) Data Scarcity}} current methods often have difficulty training to converge with limited EEG-video data.
To solve the above problems, we propose a novel framework \textbf{MindCine} to achieve high-fidelity video reconstructions on limited data.
We employ a multimodal joint learning strategy to incorporate beyond-text modalities in the training stage and leverage a pre-trained large EEG model to relieve the data scarcity issue for decoding semantic information, while a Seq2Seq model with causal attention is specifically designed for decoding perceptual information.
Extensive experiments demonstrate that our model outperforms state-of-the-art methods both qualitatively and quantitatively.
Additionally, the results underscore the effectiveness of the complementary strengths of different modalities and demonstrate that leveraging a large-scale EEG model can further enhance reconstruction performance by alleviating the challenges associated with limited data. Our Code is available at https://github.com/KevinZhou6/MindCine.
\end{abstract}
\begin{keywords}
Brain Decoding, EEG, Video Generation, Large Brain Model
\end{keywords}
\section{INTRODUCTION}
\label{sec:intro}

Decoding visual perception from brain activity is crucial to deepen the understanding of the brain's complex visual system. Spurred by the development of representation learning and generative models, numerous studies have successfully reconstructed high-quality videos from brain activity such as electroencephalography (EEG) \cite{liu2024eeg2video,liu2025mindcross,huang2025mindev,liu2025eegmirror} and functional Magnetic Resonance Imaging (fMRI) \cite{chen2023cinematic, gong2024neuroclips, lu2024animate}. In contrast to fMRI, whose temporal resolution ($\approx$0.5 Hz) is not enough to match video (24 fps), EEG signal can capture high dynamic changes due to its high temporal resolution ($\geq$200 Hz).
Hence, EEG2Video \cite{liu2024eeg2video} has been proposed to reconstruct video from EEG signals and verify that visual features and dynamic information can be decoded.

\begin{figure}[t]
    \centering
    \includegraphics[width=0.48\textwidth]{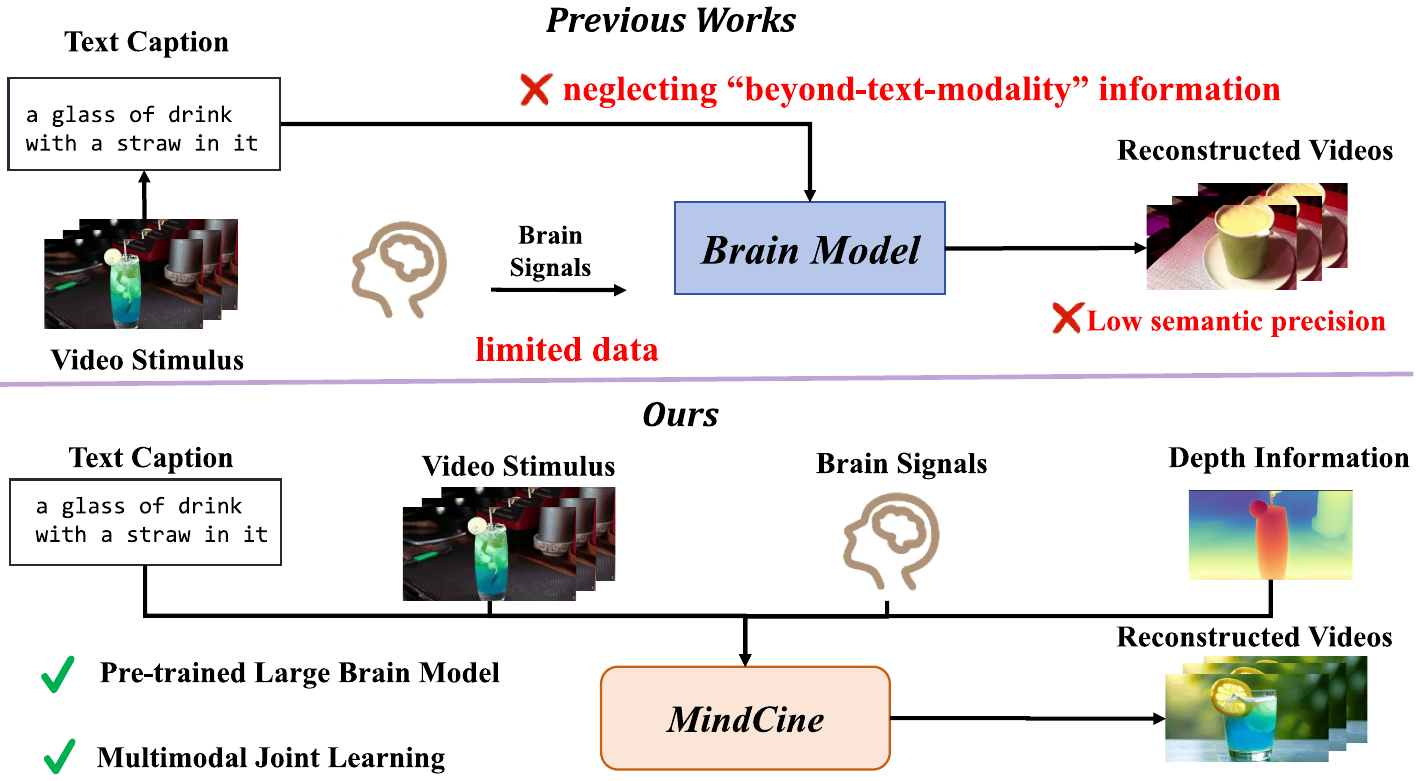}
    \vspace{-4mm}
    \caption{Brain Decoding Paradigms: Previous vs. Ours.}
    \label{fig:illus}
    \vspace{-4mm}
\end{figure}

However, EEG-to-video reconstruction still remains unsolved challenges including two aspects: \textbf{1) Single Modality} As depicted in Figure \ref{fig:illus}, EEG2Video only focuses on the alignment between EEG and text data pairs, ignoring the valuable ``beyond-text-modality" information embedded in EEG signals, resulting in the loss of multimodal information in EEG.
\textbf{2) Data Scarcity} Brain signals are considered an ``expensive" modality because they can only be obtained from real-world humans through sophisticated devices \cite{gifford2022large, gao2025cinebrain}. Consequently, the scale of available paired EEG-video datasets remains limited, posing significant challenges for effective model training.

To solve the above problems, we design a novel video reconstruction framework called MindCine, which introduces two trainable components of \textit{Semantic} and \textit{Perceptual Decoding Modules} to decode high-level and low-level information from EEG signals, respectively. \textbf{(1)} Semantic Decoding Module employs a novel multimodal joint learning strategy to realize the semantic information provided by different modalities complements each other, thus enabling the model to capture fine-grained high-level semantic information with comprehensive ``beyond-text-modality" information. \textbf{(2)} For alleviating data scarcity, we fine-tune a large-scale EEG model on the SEED-DV dataset, leveraging its strong feature learning capacity to extract generic and robust EEG representations from limited data.
\textbf{(3)} Drawing inspiration from natural language processing, the Perceptual Decoding Module introduces a Causal Sequence (CausalSeq) architecture to decode continuous low-level dynamic perceptual information from high temporal resolution EEG signals.
\textbf{(4)} During inference, MindCine utilizes a text-to-video (T2V) diffusion model, conditioned on both semantic content and low-level perceptual information, to achieve high-fidelity video reconstruction.
\begin{figure*}[h]
    \centering
    \includegraphics[width=0.9\textwidth]{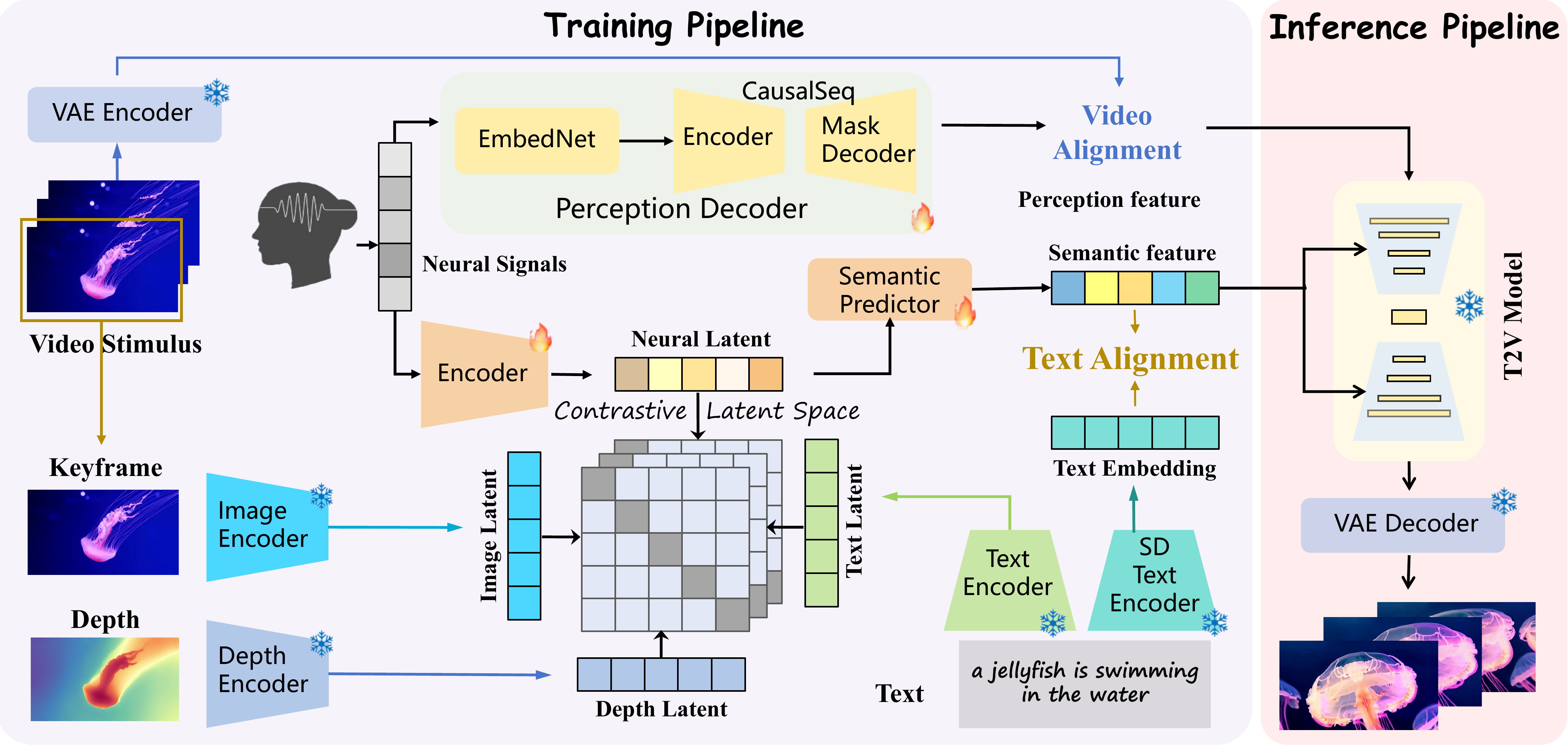}
    \caption{The overall framework of MindCine. }
    \label{fig:MindCine_framework}
\end{figure*}

In Conclusion, our contributions are as follows:

 \textbf{(1)} We propose MindCine, a novel video reconstruction framework that combines multimodal joint learning and a CausalSeq architecture to decode semantic and perceptual features from EEG.
 
 \textbf{(2)} To address the challenge of data scarcity, we introduce a large-scale EEG model and leverage its strong feature learning capability to extract generic and robust EEG representations from limited data.
 
 \textbf{(3)} Extensive experiments on the EEG-to-video benchmark demonstrate MindCine outperforms state-of-the-art methods both qualitatively and quantitatively.

\section{METHOD}
\label{sec:method}

\subsection{Semantic Decoding Module}
\subsubsection{Large-scale Pretrained Encoder}
The EEG signals are first fed into an EEG encoder to obtain neural latents for subsequent multimodal joint learning.
Usually, the EEG encoder can be implemented with simple architecture (i.e. an MLP) and trained from scratch like EEG2Video \cite{liu2024eeg2video}. However, to tackle the data scarcity in the EEG-to-video field, we exploit cutting-edge large EEG models trained based on their large-scale pretrained weights to validate whether large EEG models can benefit the video reconstruction task.
These large EEG models all follow the masked then reconstruction pretraining paradigm, like BIOT \cite{yang2023biot} and CBraMod \cite{wang2024cbramod}. EEGPT \cite{yue2024eegpt} uses an additional contrastive loss to align masked EEG data and complete EEG data, while LaBraM \cite{jiang2024large} and Gram \cite{li2025gram} apply neural quantization to learn a neural codebook for next-stage masked-predict pretraining.
\subsubsection{Multimodal Joint Learning}
Directly aligning the text captions with EEG signals to reconstruct videos will result in unsatisfactory outcomes and neglect the valuable ``beyond-text-modality" information embedded in EEG signals, since the semantics of embeddings only originate from the text modality. 
Considering the robust semantic information embedded in the latent space of CLIP \cite{lu2024animate, gao2020sketchycoco}, we adopt the SoftCLIP loss \cite{gao2024softclip} (with bidirectional component omitted for brevity) to align EEG embeddings with the latent space of CLIP:
\begin{equation}
\begin{aligned}
    \hspace{5.5em}\mathcal{L}_{\text{SoftCLIP}}(\mathbf{e}, \hat{\mathbf{e}}) =  - \sum_{k=1}^{B} \sum_{l=1}^{B}  \\ 
    & \hspace{-17em} \left[ \frac{\exp(\mathbf{e}_k \cdot \mathbf{e}_l/\tau) } {\sum_{m=1}^{B} \exp(\mathbf{e}_k \cdot \mathbf{e}_m / \tau)} \log \left( \frac{\exp(\hat{\mathbf{e}}_k \cdot \mathbf{e}_l / \tau) }{\sum_{m=1}^{B} \exp(\hat{\mathbf{e}}_k \cdot \mathbf{e}_m / \tau)} \right) \right],
\end{aligned}
\end{equation}
where $\mathbf{e}$ and $\hat{\mathbf{e}}$ are the latent representations from two modalities in a batch of size $B$. $\tau$ is a learned temperature parameter. Then, given embeddings $\mathbf{e_s}$, $\mathbf{v}$, $\mathbf{t}$, $\mathbf{d}$ from EEG, image, text, and depth encoders, respectively, the joint loss is:
\begin{equation}
\begin{aligned}
    \mathcal{L}_{\text{Joint}} = 
    \alpha_{1}\mathcal{L}_{\text{SoftCLIP}}(\mathbf{e_s}, \mathbf{v}) +  \\ \alpha_{2}\mathcal{L}_{\text{SoftCLIP}}(\mathbf{e_s}, \mathbf{t}) + &\alpha_{3}\mathcal{L}_{\text{SoftCLIP}}(\mathbf{e_s}, \mathbf{d}),
\end{aligned}
\end{equation}
these Greek letters are all hyperparameters for balancing each loss function. In this study, $\alpha_{1},\alpha_{2},\alpha_{3}$ are set to $\frac{1}{3}$.

For the sake of training stability,  we introduce an additional MSE loss to bring the EEG and text embeddings closer together:
\begin{equation}
    \mathcal{L}_{\text{Projection}} = \| \mathbf{e_s} - \mathbf{t} \|_2^2.
\end{equation}
\subsubsection{Semantic Predictor}
To condition the T2V model for reconstructing high-quality and accurate videos, we train the semantic predictor using the standard mean square error (MSE) loss function to map the semantic prediction $\mathbf{\hat{e}_t}$ to the text condition $\mathbf{e_{t}}$ of Stable Diffusion (SD), denoted as $\mathcal{L}_{Alignment}$:
\begin{equation}
\mathcal{L}_{\text{{Alignment}}} =\| \mathbf{\hat{e}_t}-\mathbf{e_{t}}\|_2^2,
\end{equation}
where $\mathbf{\hat{e}_t} = \mathcal{P}redictor(\mathbf{e_s})$, and $\mathbf{e_t}$ is acquired by feeding the caption into the frozen SD text encoder.

\subsubsection{Loss Function Co-optimization}
The loss function of the semantic decoding module consists of multiple parts, and each part plays a different role. Therefore, in order to ensure that the individual loss can work effectively and synergistically during the training process, we set the mixing coefficient $\mathcal{\lambda}$ and $\mathcal{\mu}$ to balance the multiple losses, where $\mathcal{\lambda} = 0.01$ and $\mathcal{\mu} = 0.5$: 
\begin{equation}
\mathcal{L}_{\text{Semantic}} = \mathcal{L}_{\text{Projection}} + \mathcal{\lambda}\mathcal{L}_{\text{Joint}} + \mathcal{\mu}\mathcal{L}_{\text{Alignment}}.
\end{equation}

\begin{table*}[t]
\centering
\renewcommand{\arraystretch}{0.85} 
\begin{tabular}{l|l|ccccccc}
    \toprule
    \multirow{3}[3]{*}{\rotatebox{90}{\# Classes}} & \multirow{3}[3]{*}{Method}   & \multicolumn{2}{c}{Video-based} & \multicolumn{5}{c}{Frame-based} \\
    \cmidrule(lr){3-4} \cmidrule(lr){5-9}
    ~ & &   \multicolumn{2}{c}{Semantic-Level} & \multicolumn{2}{c}{Semantic-Level} & \multicolumn{3}{c}{Pixel-Level}\\
    \cmidrule(lr){3-4}  \cmidrule(lr){5-6} \cmidrule(lr){7-9} 
    ~ & &   2-way    & 40-way  & 2-way   & 40-way & SSIM   & PSNR  & Hue-pcc  \\
    \midrule
    
    \multirow{2}{*}{10} 
        
        & EEG2Video \cite{liu2024eeg2video}     & {{0.850\scriptsize{$\pm0.02$}}} &  {0.320\scriptsize{$\pm0.04$}} &{0.798\scriptsize{$\pm0.02$}} &{0.230\scriptsize{$\pm0.02$}} & {{0.301\scriptsize{$\pm0.07$}}} & {{8.973\scriptsize{$\pm1.24$}}} & {{0.749\scriptsize{$\pm0.14$}}}  \\

        & MindCine & \textbf{0.867\scriptsize{$\pm0.03$}} & \textbf{0.331\scriptsize{$\pm0.02$}} &\textbf{{0.826\scriptsize{$\pm0.03$}}} & \textbf{0.288\scriptsize{$\pm0.02$}} & \textbf{0.333\scriptsize{$\pm0.10$}} & \textbf{9.568\scriptsize{$\pm1.42$}} & \textbf{0.775\scriptsize{$\pm0.12$}}   \\
        \midrule
   \multirow{2}{*}{40} & EEG2Video \cite{liu2024eeg2video}      &    0.800\scriptsize{$\pm0.03$}        &        {0.161\scriptsize{$\pm0.01$}}           &       0.772\scriptsize{$\pm0.03$}     &   {0.146\scriptsize{$\pm0.01$}}     &     {0.258\scriptsize{$\pm0.08$}}     &       8.684\scriptsize{$\pm1.46$} & 0.728\scriptsize{$\pm 0.08$} \\

    ~ & MindCine     & 
  \textbf{0.818\scriptsize{$\pm0.03$}} & 
  \textbf{0.179\scriptsize{$\pm0.02$}}&
    \textbf{{0.787\scriptsize{$\pm0.02$}}} & 
    \textbf{0.171\scriptsize{$\pm0.03$}} & \textbf{0.278\scriptsize{$\pm0.11$}} & \textbf{9.035\scriptsize{$\pm1.74$}} & 
    \textbf{0.768\scriptsize{$\pm0.13$}}  \\
    \midrule
        
        
        
        
        \multirow{2}{*}{40} 
        
        & w/o Semantic  & {0.797\scriptsize{$\pm0.03$}} & {0.137\scriptsize{$\pm0.02$}} & {0.736\scriptsize{$\pm0.03$}} & {0.111\scriptsize{$\pm0.01$}} & \underline{0.225\scriptsize{$\pm0.08$}} & {{7.394\scriptsize{$\pm1.48$}}} & {{0.586\scriptsize{$\pm0.12$}}} \\

        & w/o Perception & \underline{{0.808\scriptsize{$\pm0.02$}}} & \underline{0.150\scriptsize{$\pm0.04$}} & \underline{0.741\scriptsize{$\pm0.03$}} & \underline{0.149\scriptsize{$\pm0.02$}} & {0.198\scriptsize{$\pm0.09$}} & \underline{8.297\scriptsize{$\pm1.63$}}
        & \underline{0.732\scriptsize{$\pm0.10$}} \\
        
        \midrule
        


    \multirow{3}{*}{40} 
       
        & {Text} &  {{0.798\scriptsize{$\pm0.03$}}} & {0.150\scriptsize{$\pm 0.03$}}&{{0.766\scriptsize{$\pm0.03$}}} & {0.142\scriptsize{$\pm 0.02$}} &{{0.269\scriptsize{$\pm0.06$}}} & {{8.683\scriptsize{$\pm1.39$}}} & {{0.753\scriptsize{$\pm0.13$}}} \\

        & {Text+Depth}  &  {{0.804\scriptsize{$\pm0.03$}}} & {0.155\scriptsize{$\pm 0.03$}}&{{0.772\scriptsize{$\pm0.03$}}} & {0.150\scriptsize{$\pm 0.02$}} &\underline{{0.274\scriptsize{$\pm0.12$}}} & \underline{{8.899\scriptsize{$\pm1.52$}}} & \underline{{0.766\scriptsize{$\pm0.08$}}}  \\
        
        & {Text+Image}   & \underline{{0.810\scriptsize{$\pm0.04$}}} & \underline{0.166\scriptsize{$\pm0.03$}} & \underline{{0.776\scriptsize{$\pm0.04$}}} & \underline{0.155\scriptsize{$\pm0.04$}} & {{0.270\scriptsize{$\pm0.07$}}} & {{8.790\scriptsize{$\pm1.37$}}} & {{0.762\scriptsize{$\pm0.12$}}} \\

    \bottomrule
\end{tabular}
\caption{Quantitative Analysis of Video Reconstruction. Bold font signifies the best performance. In each ablation experiment, the second-best results are underlined. All metrics are calculated as the average across all subjects.}
\label{tab:cross}
\end{table*}

\subsection{Perceptual Decoding Module}
\subsubsection{EmbedNet}
Firstly, for an input two-second EEG segment $\mathbf{x} \in \mathcal{R}^{C \times T}$, where $C$ denotes the number of electrode channels and $T$ denotes the number of time samples, we apply the overlapping sliding window algorithm to slice one EEG segment into $t$ shorter EEG segments, denoted as $\mathbf{\mathcal{X}}=[x_1,x_2,...,x_{t}]$, where $t$ is the number of video frames.  Consequently, a feature extraction module named EmbedNet effectively extracts topological and spatiotemporal information from these shorter segments to extract embeddings, denoted as $\mathbf{e_p}= \{ e^1_p,e^2_p,...,e_p^t\}$. The EmbedNet is based on the Temporal-Spatial convolution architecture.

\subsubsection{CausalSeq Model}
Deriving inspiration from natural language processing (NLP), we consider each EEG embedding as a word embedding and introduce the Transformer-based CausalSeq model to decode continuous, dynamic low-level visual information from high temporal resolution EEG signals.

The proposed CausalSeq model is based on the Encoder-Decoder architecture. In our framework, the EEG embeddings and positional embeddings (PE) are jointly input into the encoder layers. In the decoder parts, the attention layer incorporates a specially designed Causal Mask to ensure that each token cannot access information from subsequent tokens. Each encoder and decoder layer contains a multi-head attention (MHA) and feed-forward network (FFN).  In each attention layers, we add layer normalization to the queries and keys before the dot-product attention mechanism to avoid over-large values in attention logits \cite{jiang2024large}. The output of the CausalSeq model is the predicted latent variables $\mathbf{\hat{z}_0}$.

As shown in Figure \ref{fig:MindCine_framework}, the Ground Truth (GT) video latent variables $\mathbf{z_0}$ are extracted by feeding video frames into the VQ-VAE encoder. Then, denoting $B$ is the batch size and $\left\|\right\|_2$ is the squared $L_2$-norm, the final perception loss is defined as:
\begin{equation}
\mathcal{L}_{\text{Perception}} = \sum_{i=1}^{B} \left\| \mathbf{\hat{z}^i_0} - \mathbf{z_0^i}\right\|_2^2.
\end{equation}
\subsection{Inference Module}  
In the inference module of MindCine, we leverage a pre-trained, off-the-shelf Text-to-Video (T2V) diffusion model as the backbone conditioned on both semantic content and perceptual information, to achieve high-fidelity video reconstruction. 

The T2V model possesses a significant amount of prior knowledge from the graphics, image, and video domains. However, since the videos generated by the diffusion model are stochastic, in the MindCine  framework, we introduce adversarial guidance to control the T2V model `what to generate' and `what not to generate' to better reconstruct the original video stimuli:
\begin{equation}
    \hat{\epsilon}_{\theta}(z_t, c, \bar{c}) = \epsilon_{\theta}(z_t, \bar{c}) + s(\epsilon_{\theta}(z_t, c) - \epsilon_{\theta}(z_t, \bar{c})),
\end{equation}
where $c$ is the positive condition, $\bar{c}$ is the negative condition, $s$ is the guidance scale and $\epsilon_{\theta}$ is the score estimator.

\section{EXPERIMENT}
\label{sec:exp}
\subsection{Datasets}
For the EEG-to-video task, our experiments are conducted on the SEED-DV dataset \cite{liu2024eeg2video}, which contains synchronized EEG-video pairs from 20 subjects. Each subject watches 7 video blocks, including 1,400 two-second video clips of 40 different concepts. In our experiments, we use the first 6 blocks (1,200 EEG-video pairs) as the training set and the last block (200 EEG-video pairs) as the test set. Subsequently, to obtain a multimodally aligned dataset, we use BLIP2 \cite{li2023blip} for textual descriptions of the images and DepthAnything v2\cite{depth_anything_v2} for depth estimation, forming an aligned text and depth dataset. 

\subsection{Evaluation Metrics}
Following prior studies \cite{liu2024eeg2video, lu2024animate}, we use the N-way-top-K metric and set K to 1 for measuring semantic-level decoding performance. We consider a video is successfully reconstructed if the GT class is among the top-K probabilities in the predicted classification results using a pre-trained classifier, selected from N randomly chosen classes that include the GT classes. The reported success rate is based on the results of 100 repeated tests. For frame-based metric (2-way-I, 40-way-I), the classifier is an ImageNet classifier \cite{radford2021learning}. For video-based metric (2-way-V, 40-way-V), the classifier is a VideoMAE-based \cite{tong2022videomae} video classifier trained on Kinetics-400 dataset \cite{kay2017kinetics}.
As for pixel-level metrics, we employ the structural similarity index measure (SSIM), peak signal-to-noise ratio (PSNR), and hue-based Pearson correlation coefficient \cite{swain1991color} (Hue-pcc) as pixel-level metrics.

 \begin{figure}[t]
    \raggedright
    \includegraphics[width=0.48\textwidth]{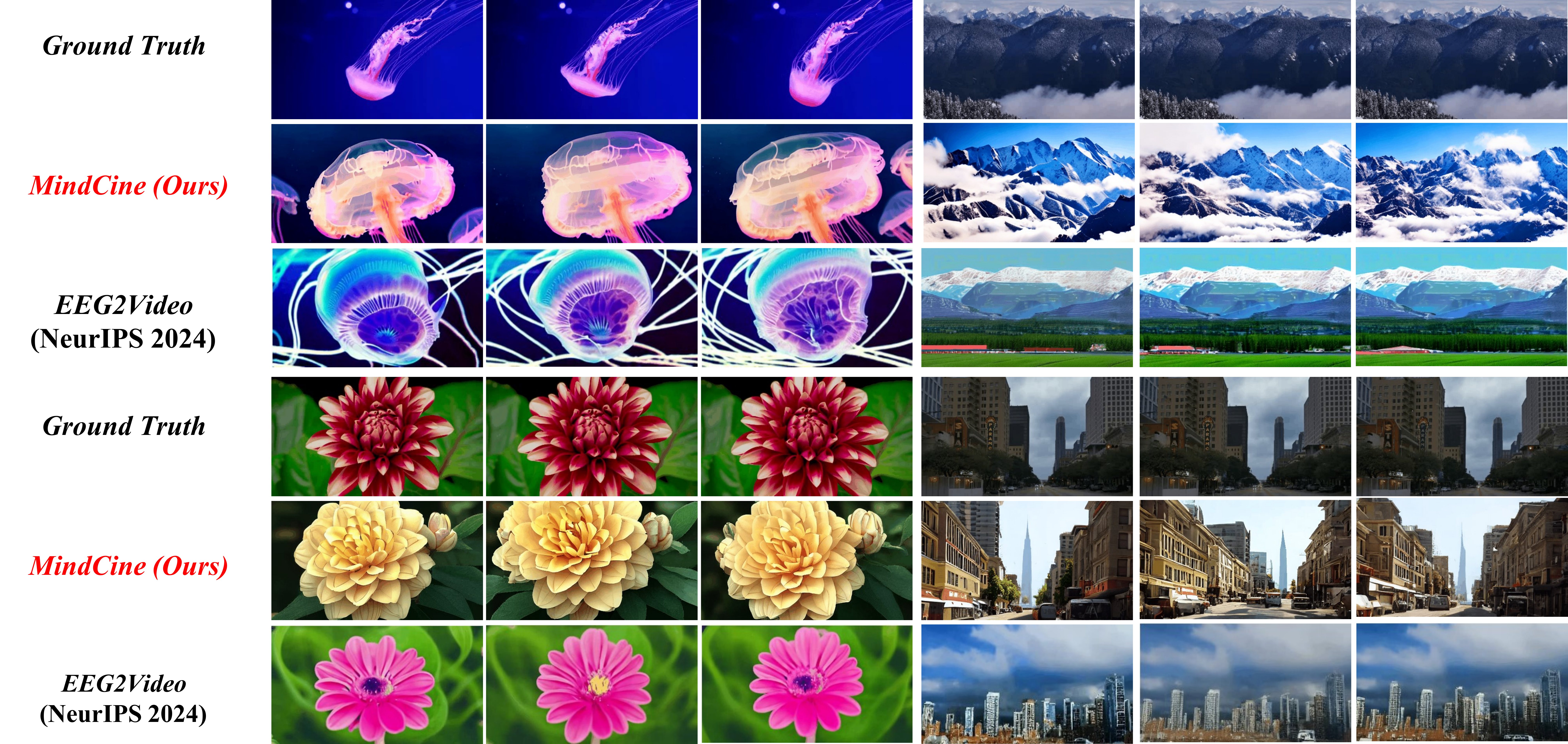}
    \vspace{-4mm}
    \caption{Successful reconstruction results on SEED-DV dataset. Our MindCine reconstructs videos with higher quality and more precise semantics.}
    \label{fig:SEED-DV}
    \vspace{-7mm}
\end{figure}

\section{RESULTS}
\subsection{Video Reconstruction}
The quantitative results for all methods are presented in Table \ref{tab:cross}, where our MindCine achieves state-of-the-art performance across all seven evaluation metrics, demonstrating its strong capability in capturing both semantic-level and pixel-level visual features. Note that here we do not use a large EEG model as the encoder for a fair comparison with EEG2Video \cite{liu2024eeg2video}, since EEG2Video is trained from scratch. Qualitative comparisons are presented in Figure \ref{fig:SEED-DV}. Our MindCine reconstructs videos with more accurate details: e.g., the flower petals are more layered, and the layout of the buildings is also more consistent with ground truth.
\begin{figure}[h]
    \centering
    \includegraphics[width=0.48\textwidth]{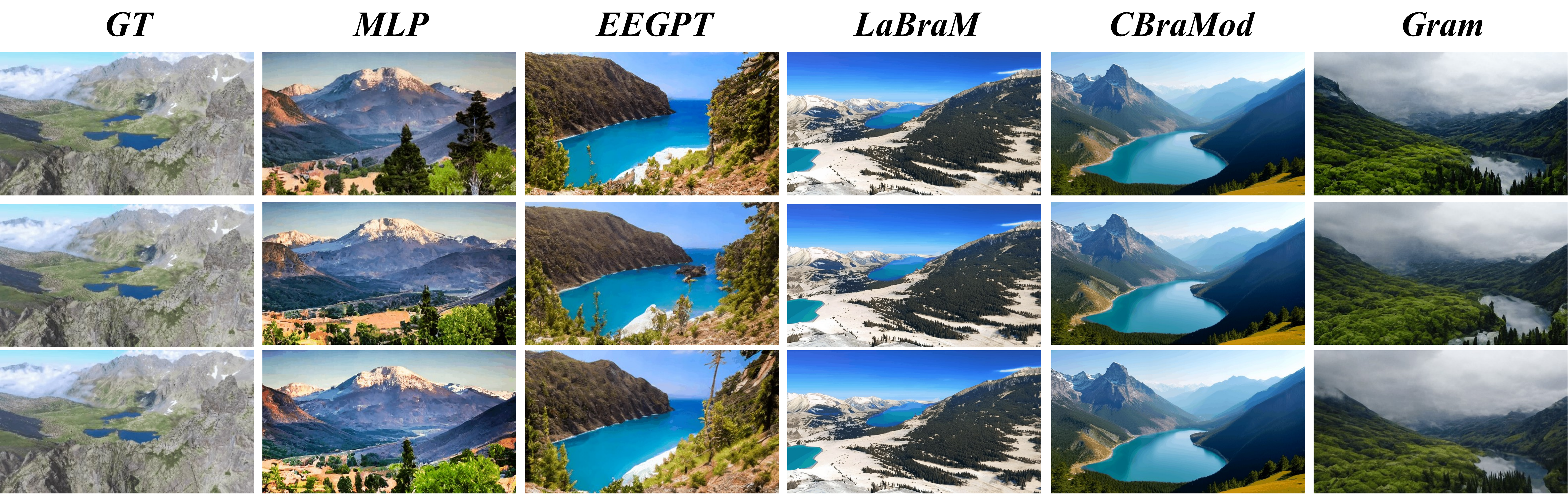}
    \vspace{-4mm}
    \caption{Successful reconstruction results with different large EEG models}
    \vspace{-5mm}
    \label{fig:llm}
\end{figure}
\subsection{Leverage Large EEG Model for Video Reconstruction} 
We select different open-source large EEG models as EEG encoders to verify their decoding ability. As illustrated in the Table \ref{tab:ELMs} and Figure \ref{fig:llm}, the use of large EEG models significantly enhances the semantic-level metrics, demonstrating that the model is able to decode richer semantic information.
Since we only adopt large EEG model in the semantic decoding module, the pixel-level metrics are not affected too much. Therefore Table \ref{tab:ELMs} only displays semantic-level metrics.
It can be observed that the largest model Gram (6.0M) outperforms other models in most metrics. But model size is not everything, like the smaller model CBraMod (4.0M) surpasses LaBraM (5.8M) in most metrics.
As seen in Figure \ref{fig:llm}, the model using a pre-trained large model as the encoder is able to capture more detailed semantic information, such as lakes between mountains. In contrast, the MLP encoder fails to decode such finer details.
\begin{table}[ht]
\footnotesize
\setlength{\tabcolsep}{2pt}

    \raggedright 
    \begin{tabular}{l c |c c c c}  
        \midrule
        \multirow{1}[3]{*}{Models} & \multirow{1}[3]{*}{Size} 
        & \multicolumn{2}{c}{Video-based} &
         \multicolumn{2}{c}{Frame-based} \\
        \cmidrule(lr){3-4} \cmidrule(lr){5-6}
        \multirow{1}{*}{} & \multirow{1}{*}{}
        & \multicolumn{1}{c}{2-way-V} 
        & \multicolumn{1}{c}{40-way-V}
        & \multicolumn{1}{c}{2-way-I} 
        & \multicolumn{1}{c}{40-way-I}\\       
        
        \midrule
        
        
       {MLP} & ------ & {{0.818\scriptsize{$\pm0.03$}}} & {0.179\scriptsize{$\pm0.02$}} & {{0.787\scriptsize{$\pm0.02$}}} & {0.171\scriptsize{$\pm0.03$}}   \\
        \cmidrule(lr){1-6}

        BIOT \cite{yang2023biot} & 3.2M& {{0.834\scriptsize{$\pm0.03$}}} & {0.227\scriptsize{$\pm0.03$}} & {{0.786\scriptsize{$\pm0.03$}}} & {0.217\scriptsize{$\pm0.04$}}   \\
        
        {LaBraM \cite{jiang2024large}} & 5.8M &{{0.842\scriptsize{$\pm0.02$}}} & {0.242\scriptsize{$\pm0.03$}} & \underline{{0.797\scriptsize{$\pm0.03$}}} & {0.224\scriptsize{$\pm0.03$}}   \\
        
        {EEGPT \cite{yue2024eegpt}} & 4.7M &{{0.835\scriptsize{$\pm0.03$}}} & {0.236\scriptsize{$\pm0.02$}} & {{0.790\scriptsize{$\pm0.03$}}} & {0.219\scriptsize{$\pm0.04$}}    \\
        {CBraMod \cite{wang2024cbramod}} & 4.0M &\underline{{0.844\scriptsize{$\pm0.02$}}} & \underline{0.245\scriptsize{$\pm0.03$}} & {{0.794\scriptsize{$\pm0.03$}}} & \textbf{0.227\scriptsize{$\pm0.02$}} \\
        
        {Gram \cite{li2025gram}} & 6.0M &\textbf{{0.847\scriptsize{$\pm0.03$}}} & \textbf{0.250\scriptsize{$\pm0.02$}} & \textbf{{0.807\scriptsize{$\pm0.03$}}} & \underline{0.225\scriptsize{$\pm0.03$}} \\
        \midrule

    \end{tabular}
    \caption{Quantitative comparison of reconstruction results on the SEED-DV dataset. The best results are highlighted in bold, and the second-best results are underlined.}
    \label{tab:ELMs}
    \vspace{-3mm}
\end{table}
\subsection{Ablation Study}

\subsubsection{Module Ablation Experiment}
As shown in Table \ref{tab:cross}, we conduct a detailed ablation study to assess the effectiveness of the two decoding modules we proposed.  After removing a certain module, the performance of the model declines significantly across nearly all metrics. The significant decline across all metrics underscores that both modules are essential components of the entire decoding framework, with each making a unique contribution to the overall performance.

\begin{figure}[ht]
    \centering
    \includegraphics[width=\linewidth]{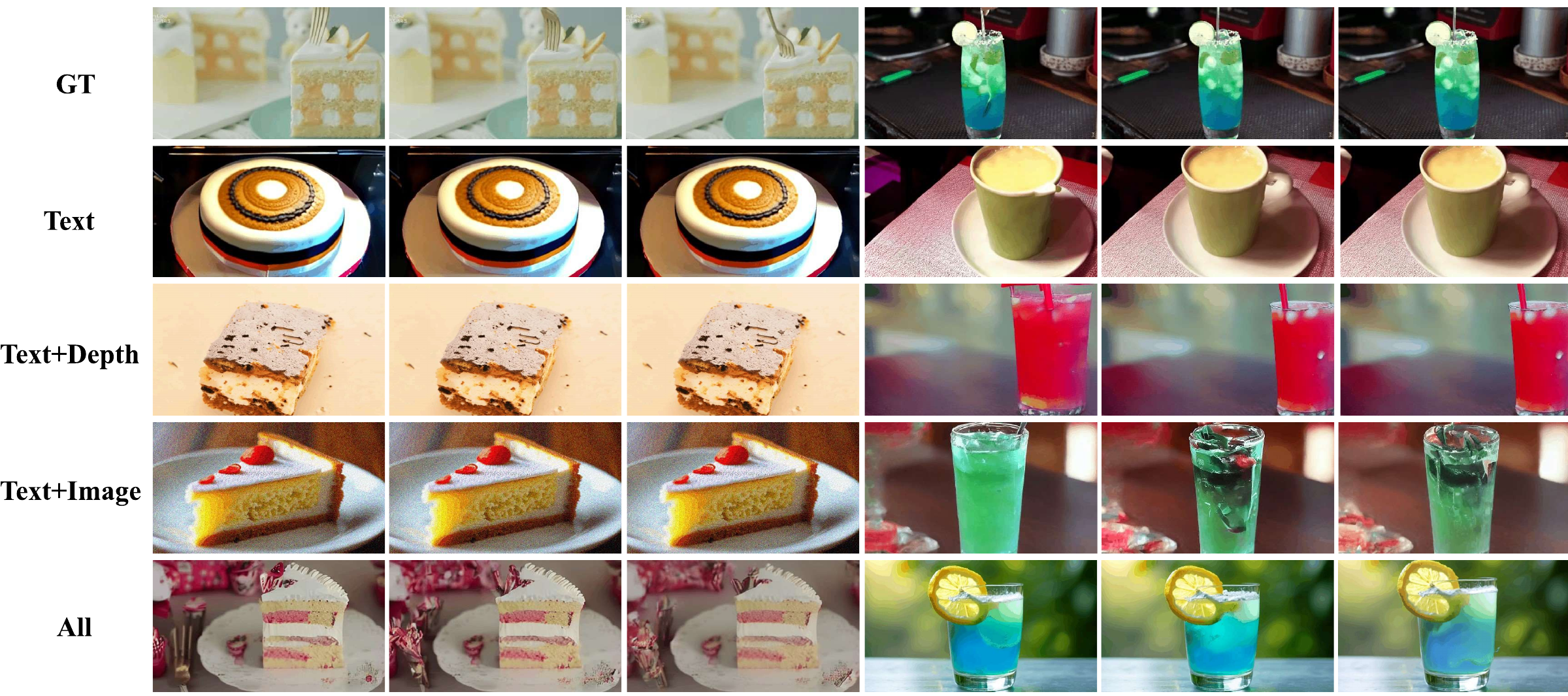}
    \caption{Successful video reconstruction results using different combinations of modalities.}
    \label{fig:modality}
    \vspace{-3mm}
\end{figure}

\subsubsection{Modality Ablation Experiment}
Table \ref{tab:cross} demonstrates that with the introduction of the image and depth modalities, the performance of the model is improved. As shown in Figure \ref{fig:modality}, there are differences in reconstruction performance when using different combinations of modalities. Due to the inherently abstract nature of textual descriptions, video reconstructions relying solely on the text modality typically lack fine-grained visual fidelity. Adding depth modality improves structural accuracy but lacks semantic richness. While combining text and images enhances semantic decoding, some fine-grained details are lost. MindCine, by jointly leveraging all three modalities, achieves more semantically coherent video reconstructions.

\section{CONCLUSION}
In this paper, we propose MindCine, a novel framework for EEG-to-video reconstruction. Our framework efficiently decodes semantic and perceptual information from EEG signals and integrates them into the latest video diffusion models to generate high-quality videos. To address the challenge of limited paired EEG-video data, we leverage a large-scale pre-trained EEG model to extract robust EEG features from limited data, significantly improving reconstruction performance. Extensive experiments on the EEG-to-video benchmark demonstrate that MindCine achieves state-of-the-art performance in video reconstruction.

\vfill\pagebreak


\section{ACKNOWLEDGEMENT}
This work was supported in part by grants from Brain Science and Brain-like Intelligence Technology-National Science and Technology Major Project (2025ZD0218900), National Key Research and Development Program of China (2024YFC3606800), National Natural Science Foundation of China (62376158), STI 2030-Major Projects+2022ZD0208500, Medical-Engineering Interdisciplinary Research Foundation of Shanghai Jiao Tong University “Jiao Tong Star” Program (YG2023ZD25, YG2024ZD25 and YG2024QNA03), Shanghai Jiao Tong University 2030 Initiative, the Lingang Laboratory (Grant No. LGL-1987), GuangCi Professorship Program of RuiJin Hospital Shanghai Jiao Tong University School of Medicine, and Shanghai Jiao Tong University SCS-Shanghai Emotionhelper Technology Co., Ltd Joint Laboratory of Affective Brain-Computer Interfaces.

\bibliographystyle{IEEEbib}
\bibliography{strings,refs}

\end{document}